\documentclass[10pt,twocolumn,letterpaper]{article}

\usepackage{cvpr}      

\usepackage{booktabs}
\usepackage{amsmath}
\usepackage{amssymb}
\usepackage{bbold}
\usepackage{pifont}
\newcommand{\cmark}{\ding{51}}%
\newcommand{\xmark}{\ding{55}}%

\newtheorem{theorem}{Theorem}

\definecolor{cvprblue}{rgb}{0.21,0.49,0.74}
\usepackage[pagebackref,breaklinks,colorlinks,allcolors=cvprblue]{hyperref}

\def\paperID{11481} 
\def\confName{CVPR}
\def\confYear{2025}

\title{Let Samples Speak: \\
Mitigating Spurious Correlation by Exploiting the Clusterness of Samples}

\author{
Weiwei Li$^{1}$\\
{\tt\small davelee.uestc@gmail.com}
\and
Junzhuo Liu$^{1}$\\
{\tt\small junzhuo.cs@gmail.com}
\and
Yuanyuan Ren$^{2}$ \\
{\tt\small yuanyuanren043@outlook.com}
\and
Yuchen Zheng$^{2}$\\
{\tt\small ouczyc@outlook.com}
\and
Yahao Liu$^{1}$\\
{\tt\small lyhaolive@gmail.com}
\and
Wen Li$^{1}$\\
{\tt\small liwenbnu@gmail.com}
\and
$^{1}$University of Electronic Science and Technology of China \\
$^{2}$Shihezi University 
}

\begin{document}
\maketitle
\begin{abstract}
Deep learning models are known to often learn features that spuriously correlate with the class label during training but are irrelevant to the prediction task. Existing methods typically address this issue by annotating potential spurious attributes, or filtering spurious features based on some empirical assumptions (\eg, simplicity of bias). However, these methods may yield unsatisfactory performance due to the intricate and elusive nature of spurious correlations in real-world data. In this paper, we propose a data-oriented approach\footnote{Codes and checkpoints are available at \url{https://github.com/davelee-uestc/nsf_debiasing}.} to mitigate the spurious correlation in deep learning models. We observe that samples that are influenced by spurious features tend to exhibit a dispersed distribution in the learned feature space. This allows us to \emph{identify} the presence of spurious features. Subsequently, we obtain a bias-invariant representation by \emph{neutralizing} the spurious features based on a simple grouping strategy. Then, we learn a feature transformation to \emph{eliminate} the spurious
features by aligning with this bias-invariant representation. Finally, we \emph{update} the classifier by incorporating the learned feature transformation and obtain an unbiased model. By integrating the aforementioned identifying, neutralizing, eliminating and updating procedures, we build an effective pipeline for mitigating spurious correlation. Experiments on image and NLP debiasing benchmarks show an improvement in worst group accuracy of more than 20\% compared to standard empirical risk minimization (ERM). 
\end{abstract}    
\section{Introduction}
\label{sec:intro}

\begin{figure}[!t]
    \centering
    \includegraphics[width=\linewidth]{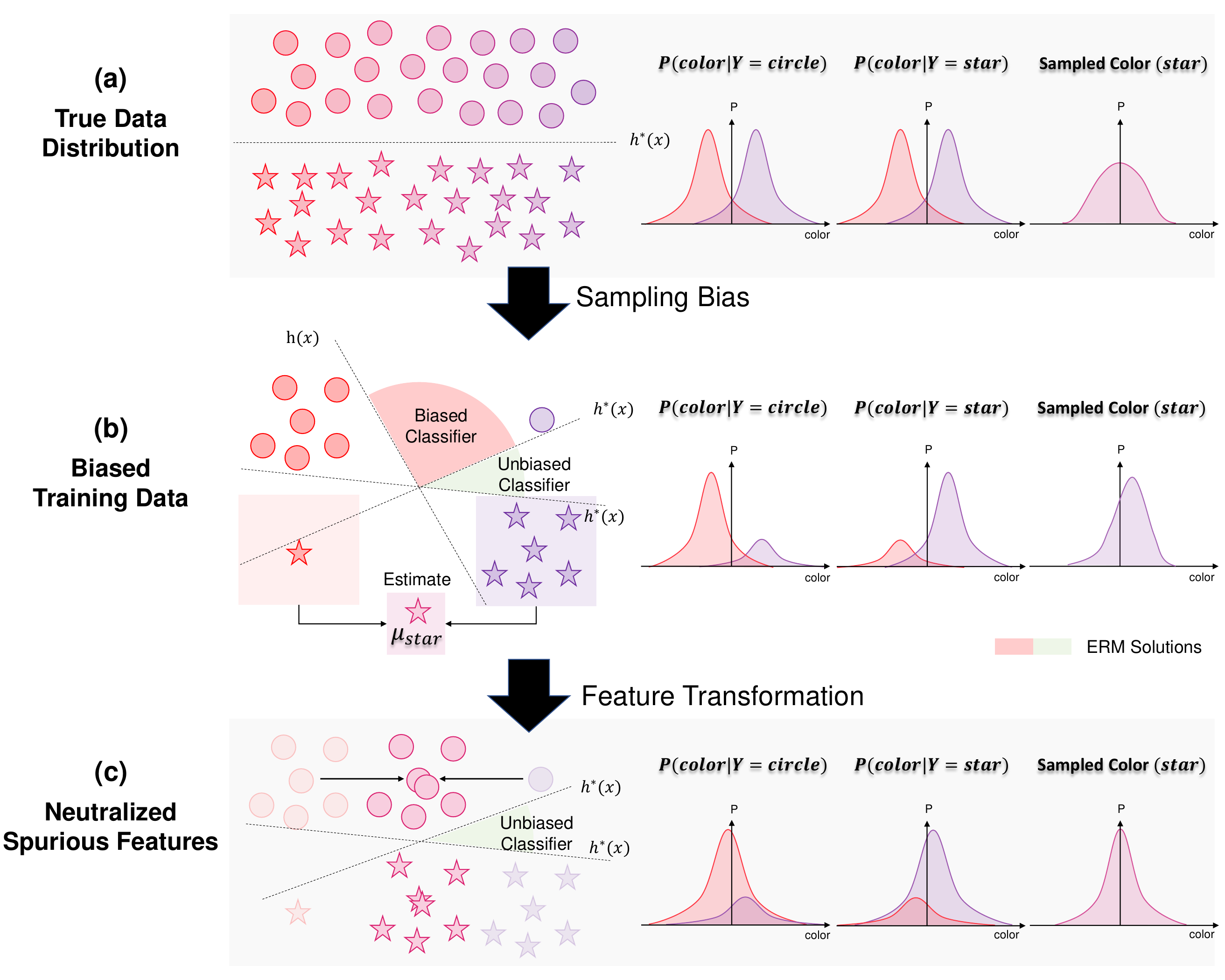}
    \caption{1) Ideally, bias attributes (\eg, color) should be evenly distributed and non-predictive of the class; 2) Sampling bias can introduce unintended patterns, like most circles being red and most stars being purple, causing some features to mistakenly correlate with class labels. Since ERM training minimizes the mean loss, an ERM-trained model is highly likely to fit these spurious correlations due to their large population in the data; 3) Intuitively, a transformation producing invariant representation for different values of bias attributes reduces the possible of learning bias.
    }
    \label{fig:fig0}
\end{figure}
Recent studies reveal that deep neural networks (DNNs) learn unintended decision rules from spurious correlations \cite{geirhos2020shortcut}, also known as model bias. For example, researchers \cite{bissoto2019constructing,rieger2020interpretations,goelmodel,zech2018variable,oakden2020hidden} report that models infer disease using cues of medical devices rather than symptoms. The sampling bias is blamed for introducing spurious correlations between attributes and class labels in data \cite{torralba2011unbiased,beery2018recognition,an2021why}, as illustrated in \cref{fig:fig0}. However, such unexpected biases are often masked by satisfying performance on i.i.d. test data \cite{seo2022unsupervised,de2024mitigating}, and hard to annotate in advance, making methods \cite{sagawa2019distributionally,kirichenko2022last,tsirigotis2024group} requiring bias attributes or labels less practical.

One possible way of addressing unknown model bias is to eliminate bias-relating features (known as spurious features) to obtain a bias-invariant representation so that the classifier does not rely on them. The key challenge of such methods is to make proper assumptions of possible spurious features. Some methods \cite{li2023whac,tiwari2023overcoming} use assumptions from the empirical manifestation of biases such as assuming simpler features, like color, to be spurious. Although seemingly straightforward, this assumption is task-specific, and has a clear limitation: they fail if the task changes, for example, changing from classifying numbers to colors in the Colored-MNIST dataset \cite{arjovsky2019invariant}. 

Instead, we estimate the bias-invariant representation using the fact that samples affected by spurious features tend to exhibit a dispersed distribution in the feature space. Specifically, we propose a task-independent assumption of spurious features, termed the strong spurious assumption, to estimate the bias-invariant representation without bias attributes or labels. This assumption is supported by recent findings: while DNNs produce both core and spurious features \cite{hermannfoundations}, DNNs favor strong features (high availability), regardless of whether they are core (predictive) or spurious (bias-related) \cite{kirichenko2022last,sagawa2020investigation}, meaning that the model becomes biased only when spurious features are stronger. This strong spurious assumption defines a characteristic in feature space for same class samples with different spurious feature values: the samples of majority values are closer to their centroid while the minority deviates. As illustrated in \cref{fig:fig0} (b), when the bias is most circles are red while most stars are purple, stronger spurious feature (color) makes purple circles closer to star class centroid rather than circle class centroid while the red circles don't, making samples of different spurious attribute separable. This characteristic enables us to (1) use the presence of a minority sample (purple circles) as the indicator of spurious features; (2) distinguish the majority (red circles) and minority groups (purple circles) and estimate a bias-invariant representation using the groups found. We provide theoretical proof using a widely accepted debiasing data model \cite{sagawa2020investigation} for binary classification. 

Building on this foundation, we introduce Neutralizing Spurious Features (NSF), a debiasing method that does not require prior knowledge of bias attributes. NSF consists of four key steps:
\textit{(1) Identifying Bias Presence}: Minority samples that deviate from the class centroid are identified, as such deviations indicate the presence of spurious features.
\textit{(2) Neutralizing Spurious Feature for Bias-Invariant Features}: Use identified groups to estimate a bias-invariant representation for each class.
\textit{(3) Eliminating Spurious Feature}: Learn a common transformation across all classes that aligns all training samples within a class to the estimated bias-invariant features. This transformation eliminates spurious features while preserving core features.
\textit{(4) Updating Classifier}: Finetune the classifier on these bias-invariant features, forcing reliance on core features alone.

To validate the effectiveness of the proposed method, we conduct experiments on multiple popular benchmarks. Experiments across four image and text tasks with known spurious correlations and one medical dataset show an average improvement of 20\% in Worst Group Accuracy (WGA) compared to Standard training via ERM, achieving state-of-the-art with a very fast speed (within a few minutes). We performed ablation studies and qualitative analysis to validate the key components and intuition.

The contributions of this paper are as follows.
\begin{itemize}
    \item We leverage the separable of samples affected by spurious features in feature space, enabling an estimation method for bias-invariant representation. We provide theoretical proof of its correctness.
    
    \item We introduce a novel, non-intrusive debiasing framework with four steps—bias identification, neutralization, spurious feature elimination, and classifier update—enabling robust learning on core features without bias labels.
    \item Extensive experiments across multiple challenging benchmarks and comprehensive ablation studies validate our approach, demonstrating its state-of-the-art performance, effectiveness of individual components.
\end{itemize}

\section{Related Works}
\label{sec:rw}
\textbf{Removing Known Spurious Correlations} Spurious correlations are common in real-world datasets \cite{geirhos2020shortcut} due to natural relationships \cite{geirhos2020shortcut} or selection bias \cite{torralba2011unbiased}. A common assumption is that bias attributes are known and labeled. This ideal assumption is the basis of the group-robustness methods \cite{sagawa2019distributionally} and retraining-based methods \cite{kirichenko2022last,chen2024understanding,deng2024robust}, which are the upper bound of debiasing methods, despite the impractical of labeling bias. A more reasonable assumption is that only the bias attributes are known, and we do not have access to the bias label. 
The task-specified biases mostly explored are that low-level features \cite{MishraSBACS0022,bahng2020learning,nam2020learning,kim2020learning,chuah2022itsa,chuah2023information,he2023shift,kim2021biaswap,geirhos2018imagenet} such as the color and texture, and background shortcut \cite{li2023whac,sauer2021counterfactual}, these biases are common in image classification. Additionally, some work \cite{adila2024zero} uses zero-shot insights from language models to obtain priors of harmful biases. Leveraging these priors of specific biases, methods such as data augmentation, can be designed and applied to address specific biases.

\textbf{Removing Unknown Spurious Correlations} Biases vary across tasks and manifest as unexpected patterns, making unknown biases a more common problem. As we know nothing about the bias, an assumption of bias is required. 
One representative assumption is the simplicity bias \cite{tiwari2023overcoming,teney2022evading}, which refers to features that are easy-to-learn or learned at the early training phase are more likely to be a bias. 
It is also common to assume that an ERM-trained model is biased and leverages the learned bias for debiasing. Methods such as Lff \cite{nam2020learning} and Echoes \cite{hu2023echoes} learn a model that differs from the biased model by minimizing the mutual alignment of the two. It is also natural to extend this idea to learning a set of diverse hypotheses, as seen in \cite{lee2023diversify,teney2022evading}. Furthermore, some works \cite{lim2023biasadv,chuah2022itsa} use bias-adversarial augmentation for model debiasing. Digging bias-conflicting samples is another solution for addressing unknown biases, considering that the spurious correlations could be effectively eliminated if the effect of the minority groups, known as the bias-conflicting samples, were amplified. A straightforward idea is reweighting those bias-conflicting samples \cite{seo2022unsupervised,nam2022spread}, or reforming the representation space using those samples \cite{zhang2022correct,hong2021unbiased,yenamandra2023facts}. However, those methods usually rely on an empirical set of bias-conflicting samples such as false positives, and require to retrain the model. The limitations of these methods arise because these assumptions use indirect manifestations of the biases and do not capture the intrinsic nature of the biases, thus being costly in training and yielding unsatisfactory performance. In contrast, the proposed method directly eliminates the spurious features by adding a single linear transformation after the frozen encoder, leveraging a derived conclusion from the bias-fitting mechanism.

\section{Problem Statement}
\label{sec:preliminaries}
We start by defining our problem setting. Formally, let $\mathcal{D} = \{(\vec x_i, y_i, g_i)\}_{i=1}^n$ denote the dataset, where \(\vec x_i \) is the input feature, \( y_i \) is the corresponding label, and $g_i = (a_i, y_i) \in G$ is the corresponding group defined by the label $y$ and a spurious attribute $a \in A$ that spuriously correlates with the label (i.e., $G = A \times Y$). 

\subsection{Masked Poor Generalization on Minorities} The ERM is a common practice of training DNNs, which aims to find the hypothesis \( h^* \in \mathcal{H} \) that minimizes the empirical risk under the loss function \(\ell : \mathcal{H} \times \mathcal{X} \times \mathcal{Y} \to \mathbb{R} \):

\begin{equation}
    h^* = \arg \min_{h \in \mathcal{H}} \frac{1}{n} \sum_{i=1}^n \ell(h(\vec x_i), y_i),
\end{equation}

where \( \mathcal{H} \) is the hypothesis class, \( \mathcal{X} \) and \( \mathcal{Y} \) denote the input and output spaces respectively. The ERM selects a model from a hypothesis class that minimizes average loss over training data, and results in majority groups having a greater impact, 
so the ERM makes a good average performance, but underperforms for minority groups \cite{tsirigotis2024group}.

\textbf{Learning Goal: Optimizing Worst Group Accuracy} Our goal is to learn a model \( h : \mathcal{X} \to \mathcal{Y} \) that maximizes the least accurately predicted subgroup, also known as Worst Group Accuracy (WGA). This goal ensures that the performance of minority groups is taken into account. The WGA is defined as:
\begin{equation}
    \text{Acc}_{wg}(h) = \min_{j \in \{1, 2, \ldots, k\}} \frac{1}{|\mathcal{D}^{j}|} \sum_{(\vec x_i, y_i) \in \mathcal{D}^{j}} \mathbb{1}[h(\vec x_i) = y_i],
\end{equation}

where $\mathcal{D}^j$ denote $j$-th group, \( \mathbb{1}[\cdot] \) is the indicator function, $h$ is the classifier.

\subsection{The Challenge of Unknown Biases} Considering the ERM objective, poor generalization on minority groups indicates that the model learns patterns aligned with the majority group but not consistently true within each class. These spurious correlations—valid for the majority but not the minority—lead models to make inaccurate predictions for minority groups \cite{geirhos2020shortcut}.

If the groups within the dataset are known, we can address the issue of poor performance on minority groups by using group reweighting or resampling such as \cite{sagawa2019distributionally,kirichenko2022last}. These techniques ensure that the model pays equal attention to all groups, thereby improving performance on the least accurately predicted subgroup. However, in many real-world scenarios, the groups are not explicitly labeled, and how the groups formed is unknown, making it hard to improve the performance of the model on minorities. 

\begin{figure*}[!t]
    \centering
    \includegraphics[width=\linewidth]{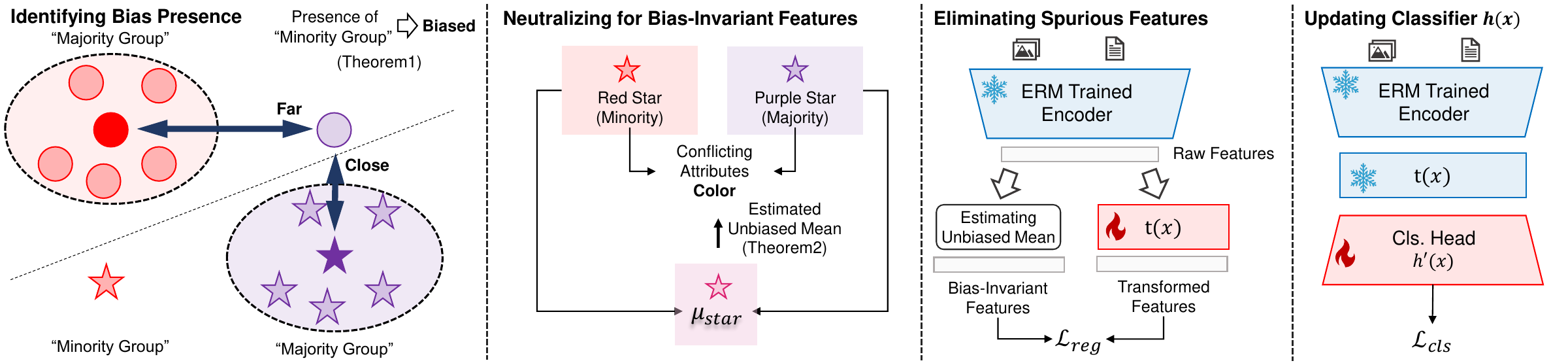}
    \caption{NSF leverages a task-independent strong spurious assumption, enabling us to (1) use the presence of minority sample (purple circles) as the indicator of spurious features; (2) distinguish majority (purple stars) and minority groups (red stars) and estimate a bias-invariant representation using the groups found. NSF mitigates biases by (3) first eliminating the bias attributes by transforming features to align training samples with estimated unbiased mean values of the true data distribution, then (4) debiasing classifiers through fine-tuning.}
    \label{fig:pipe}
\end{figure*}
In summary:
\begin{itemize}
    \item ERM minimizes average loss over the dataset but often overfits to spurious features prevalent in majority group data, masking poor generalization on minority groups.
    \item Models fitting spurious correlations cause incorrect predictions for minority groups. However, in many real-world scenarios, the groups are not explicitly labeled, and how the groups formed is unknown, making it hard to apply group reweighting or resampling techniques.
\end{itemize}

\section{Proposed Method}

\subsection{Overview}
To address the model bias, especially considering the bias attributes are hard to foresee before the model is trained, we propose NSF, a novel debiasing method that does not require prior knowledge of bias attributes. We develop a method to estimate mean values of the true data distribution without accessing the bias labels. Thus, NSF can eliminate the spurious features by transforming features to align training samples with estimated unbiased mean values of the true data distribution, and debias the classifier through fine-tuning, as in \cref{fig:pipe}.

\subsection{The Biased-Sampled Data Model} 
To address these spurious correlations, it is essential to understand their origins. Here, we use the widely used \cite{sagawa2020investigation,deng2024robust,chen2024understanding} assumption of sampling bias. It refers to certain members of a population being systematically more likely to be selected than others, leading to non-representative samples \cite{an2021why}. 
Sampling bias can cause labels to mistakenly correlate with a specific attribute because the samples are not representative of the entire population, as in \cref{fig:fig0}. 
We adopt a data generation process from \cite{sagawa2020investigation} to model the joint data distribution $(X_{\rho},Y_{\rho},A_{\rho}) \sim p_{\rho}$ under spurious correlation. The label $y \in Y_{\rho}$ follows the Uniform distribution over $\{1, -1\}$, the data point $\vec x=[Ba, y,\vec \delta] \in X_\rho$ and the spurious feature $a \in A_\rho$  are generated as follow:
\[
 a \sim  
    \begin{cases}
    P(a=k|y=k)=\rho & \\
    P(a=-k|y=k)=1-\rho & 
    \end{cases}, \vec \delta \sim \mathcal{N}(\vec 0,\vec I^{D-2}),
\]

where $\mathcal{N}$ is the normal distribution, D is the dimension of $\vec x$, $\rho \in (0.5, 1) $ , and $B \geq 1$  is scalar constants. And The conditional expectation of $\vec x$ is 
\begin{equation}
    \label{equ:cm}
    \mathbb{E}(\vec x|y=k) =  [(2\rho-1)Bk,k,\vec 0].
\end{equation}
This data generation process models the characteristics of the bias-sampled data, if $\rho=0.5$, then it means that the sampling is fair, $a$ has no correlation with the label, or else the sampling procedure is biased toward a specific value. This creates a majority group with a high concentration of a particular value, while minority groups, known as bias-conflicting samples, form a small population \cite{torralba2011unbiased}. In this data model, the input $\vec x$ consists of core feature $y$, spurious feature $a$ and noise $\vec \delta$: 
\begin{itemize}
    \item \textbf{Spurious Feature $Ba$} : The spurious feature $a$ correlate with the label $y$ with a probability of $\rho$. The scalar constant $B$ controls the impact of the bias attribute $a$. The bias-sampled data has a biased conditional expectation $(2\rho-1)Bk$ for the spurious feature, and it is zero for the true data distribution ($\rho=0.5$). 
    \item \textbf{Core Feature $y$} : The core features of samples only relate to their class labels, so the label $y$ is used as the core feature since the informative is equivalent. Its conditional expectation is independent of $\rho$ and $B$.
   \item \textbf{Noise $\vec \delta$} : Other features that not correlate with the labels. Its conditional expectation is independent of $\rho$ and $B$.
\end{itemize}

So we can find that the model fits a data distribution that deviates from the true data distribution due to the sampling bias, more specifically, in the spurious feature. 

\subsection{Confirming Bias Presence}
\label{sec:susf}
In this section, we demonstrate that if the spurious features are sufficiently strong, the presence of bias can be confirmed using the relative distance to the class centroid. We start with the definition of $C_{k}^{\rho}$ and the relative distance $d(\vec x_i)$. The conditional mean $C_{k}^{ \rho} = \mathbb{E}[X_\rho | Y=k] $, also known as the centroid, can be estimated as
\begin{equation}
C_{k}^{\rho}= \frac{1}{\sum_{i=1}^{N} \mathbb{1}[\vec y_i = k]} \sum_{i=1}^{N} \mathbb{1}[\vec y_i = k] * \vec x_{i},
\end{equation}

where $\vec x_i$ is the feature. $\forall{(\vec x_i, y_i) \in p_{\rho}}$, the relative distance between $\vec x_i$ and its corresponding centroid $C_{y_i, \rho}$, compared to the nearest centroid of another class is given by $d(\vec x_i,\rho)= (\vec x_i-C_{y_i}^{\rho})^2-(\vec x_i-C_{q_i}^{ \rho})^2$, where $\vec q_i=\arg \min_{u \neq y_i} \{ (\vec x_i-C_{u}^{\rho})^2\}$. 

Based on the data model, we found that if some samples significantly deviate from the mean values, indicating the existence of spurious features, these samples are in the minority group. This assumption is natural for the bias-sampled data, taking the example of classifying circles v.s. stars, the color is a strong spurious feature. However, this spurious correlation isn't valid for all samples due to exceptions like purple circles, noticing the bias is that most circles are red. Using this conclusion, we can separate the circles of different spurious features (red and purple) in the feature space. 
Based on this insight, we propose a method for separating the minority group from the majority group by the relative distance $d(\vec x_i,\rho)$ between data points and sample means, as

\begin{theorem} If $1-(2\rho-1)^2B^4<0$, then 
\label{theorem:soft-assign}
\[
    \forall{\vec y_i=\vec y_j}, d(\vec x_i,\rho) \times d(\vec x_j,\rho) < 0 \iff a_i \neq a_j 
\]
\end{theorem}

For detailed proof, please refer to the Appendix. Theorem \ref{theorem:soft-assign} implies that we can confirm the existence of spurious feature \((a_i\neq a_j)\) by checking if any samples are deviating from their conditioned sample mean (pair of instance $i, j$ satisfying \(d(\vec x_i,\rho) \times d(\vec x_j,\rho) < 0\)).

\subsection{Estimating Bias-Invariant Representation}
To mitigate the impact of spurious features, a feasible solution is to neutralize them to obtain a bias-invariant feature, which requires estimating the unbiased mean of spurious features depending on the inaccessible bias label $a$. However, Theorem \ref{theorem:soft-assign} implies that, if the spurious feature $a$ is strong enough, we can separate the minority group from the majority group using the sign of relative distance $d(\vec x_i)$, without knowing the exact value of $a$. This enables us to estimate the unbiased conditional mean $C_k=C_{k}^{ 0.5}$ of $\vec x$ in the true data distribution. We start by identifying the majority and minority groups in each class.

\textbf{Identifying Majority and Minority Groups} We make a soft-assignment $q_i \in Q$ for each point as $q_i = \arg \min_u{||\vec x_i - C_{u}^{\rho}||_{2}}.$ And it has $y_i=q_i \Rightarrow d(\vec x_i,\rho)<0,$ and $y_i\neq q_i \Rightarrow d(\vec x_i,\rho)>0.$ Then, we can split class $X_k$ into $U_k=\{\vec x_i \mid q_i \neq y_i, (\vec x_i, y_i) \in p_{\rho}\}$ and $V_k=\{\vec x_i \mid q_i = y_i, (\vec x_i, y_i) \in p_{\rho}\}$ by $Q$. 
Since not all class k satisfies that $|U_k| > 0$ and $|V_k| > 0$ (when the spurious features not strong enough), which indicates that we cannot estimate $C_k$ for those classes. We exclude those classes with a mask $o\in O$ as $o_i=(|U_{y_i}| > 0) \land (|V_{y_i}| > 0).$ 
\textbf{Estimating Bias-Invariant Representation}
Using the groups formed by the sign of relative distance $d(\vec x_i,\rho)$, we can estimate the value of $C_{k}$ as follows.
\begin{theorem}
\label{theorem:unbias-esti}
If $1-(2\rho-1)^2B^4<0$, then 
\[
C_{k}=\mathbb{E}(\frac{1}{2|U_k|}\sum_i^{|U_k|} \vec u_i +\frac{1}{2|V_k|}\sum_j^{|V_k|} \vec v_j)
\]
where $U_k=\{\vec x \mid (\vec x, y) \in p_{\rho}, y=k, d(\vec x,\rho)>0\}$, $V_k=\{\vec x \mid (\vec x, y) \in p_{\rho}, y=k\} \setminus U_k$, $\vec u_i \in U_k$,  $\vec v_i \in V_k$.
\end{theorem}
For detailed proof, please refer to the Appendix.
\subsection{Eliminating Unknown Spurious Features}
\begin{figure}[t]
    \centering
    \includegraphics[width=\linewidth]{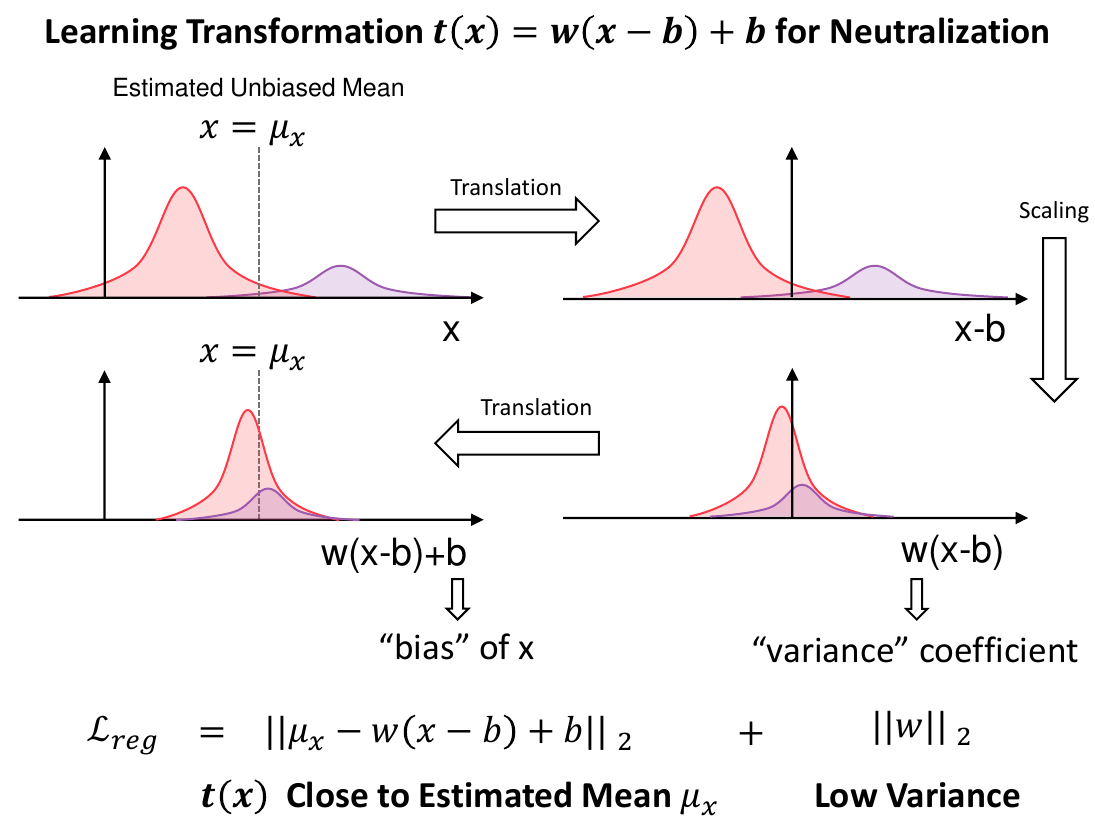}
    \caption{A linear transformation t(x) is learned to eliminate the spurious features by shifting them to their unbiased mean and reducing variance so that the correlation between the spurious feature and label is removed.}
    \label{fig:tx}
\end{figure}

Using the estimated unbiased mean \(C_y = u(a, y, \vec \delta) \) of the true data distribution, spurious features can be eliminated, as in \cref{fig:tx}, by learning a channel-wise transformation \(t(\vec x)=\vec w(\vec x- \vec b)+\vec b\) where $\vec w \in R^{1\times D}$ and $\vec b \in R^{1\times D}$ 
to make all data points close to their corresponding conditioned mean value. 

We here provide a simple verification that the corresponding channels of core features are kept unchanged, while those of spurious features are eliminated after transformation. Using \cref{equ:cm} (with \(\rho=0.5\) in the true data distribution), we have \(C_y = u(a, y, \vec \delta) = [0,y,\vec 0], x=[Ba,y,\delta]\), and we learn a \(t(\vec x)=\vec w(\vec x- \vec b)+\vec b=[w_1(Ba- b_1), w_2(y-b_2),\vec w_3(\vec \delta- \vec b_3)]\) to make \(t(x)=C_y\), then 
\begin{itemize}
    \item Spurious:  $w_1(Ba- b_1)=0\xleftarrow[]{\tiny optimal}w_1=0, b_1=0$
    \item Core: $w_2(y- b_2)=y\xleftarrow[]{optimal}w_2=1, b_2=0$
    \item Noise: $\vec w_3(\vec \delta- \vec b_3)=\vec 0\xleftarrow[]{optimal}w_3=0, b_3=0$
\end{itemize}
So optimal \(t(x)=[0,y,\vec 0]\), making core features kept unchanged, and spurious features and noise are eliminated after the transformation.

\textbf{Learning the Feature Transformation t(x)} The optimization objective is to minimize the Euclidean distance between the unbiased mean and the transformed data points
\begin{equation}
    \mathcal{L}_{reg} = \lambda ||\vec w||_2 + \frac{1}{N}\sum_i^N o_i ||t(\mathbf{sg}[\vec x_i]) - {C_{y_i}}||_{2} , 
    \label{equ:reg}
\end{equation}
where \(\mathbf{sg[\cdot]}\) is the stop-gradient operation. This objective ensures that the core feature remains unchanged, while the impact of spurious features and noise is reduced.

    \begin{table*}[!t]
    \centering
    \setlength{\tabcolsep}{1mm}
    \small{
    \begin{tabular}{lccccccccccc}
    \toprule
    &\multicolumn{2}{c}{\textsc{Labels}} & \multicolumn{2}{c}{\textsc{Waterbirds}} & \multicolumn{2}{c}{\textsc{CelebA}} & \multicolumn{2}{c}{\textsc{MultiNLI}} & \multicolumn{2}{c}{\textsc{CivilComments}} & \textsc{Mean} \\
    \cmidrule(r{4pt}){2-3} \cmidrule(r{4pt}){4-5} \cmidrule(r{4pt}){6-7} \cmidrule(r{4pt}){8-9} \cmidrule(r{4pt}){10-11} \cmidrule(r{4pt}){12-12}
    & \textsc{Tr.} & \textsc{Val} & \textsc{i.i.d.} & WGA & \textsc{i.i.d.} & WGA& \textsc{i.i.d.} & WGA & \textsc{i.i.d.} & WGA  & WGA \\ 
    \midrule
    
    \textsc{ERM} & \xmark & \xmark & $97.30$ & $72.60$ & $95.60$ & $47.20$ & $82.09$ & $68.11$ & $92.34$ & $57.06$&61.24\\
    \midrule
     \textsc{JTT\cite{liu2021just}}& \xmark & \xmark &93.30&86.70&{88.00}& 81.10& $78.60$ & 72.60 & {91.10} & 69.30 &77.43 \\    
          \textsc{MT\cite{asgari2022masktune}}& \xmark & \xmark & $93.00$ & $86.40$ & $91.30$ & $78.00$ &\multicolumn{5}{c}{------------ Not Applicable ------------} \\

\textsc{CNC\cite{zhang2022correct}} & \xmark & \xmark &90.90&{88.50}&{89.90}& \textbf{88.80}& - & - & 81.70 & 68.90 & - \\
\textsc{AFR\cite{qiu2023simple}}& \xmark & \xmark &94.20&{90.40}&{91.30}& {82.00}& {81.40} & \textbf{73.40} & 89.80 & 68.70  & 78.63 \\

      \textsc{\textbf{Ours}}  & \xmark & \xmark & 95.65\tiny{$\pm$ 0.0011} & $\textbf{91.12}$\tiny{$\pm$ 0.0063} & ${88.70}$\tiny{$\pm$ 0.0036} & ${84.27}$\tiny{$\pm$ 0.0047} & ${80.43}$\tiny{$\pm$ 0.0003} & ${73.12}$\tiny{$\pm$ 0.0008} & $87.19$\tiny{$\pm$ 0.0010} & $\textbf{79.51}$\tiny{$\pm$ 0.0022} &\textbf{82.01} \\
      \midrule
       \textsc{GDRO\cite{sagawa2019distributionally}}& \cmark & \cmark & $93.50$ & $91.40$ & $92.90$ & $88.90$ & 81.40& 77.70& 88.90& 69.90 &81.98\\
       \textsc{DFR\cite{kirichenko2022last}} & \xmark & \cmark &94.20&92.90&91.30& 88.30& 82.10 & 74.70 & 87.20 & 70.10 & 81.50\\
              \textsc{uLA\cite{tsirigotis2024group}} & \xmark & \cmark & $91.50$ & $86.10$ & ${93.90}$ & $86.50$&\multicolumn{5}{c}{------------ Not Applicable ------------}\\
    \bottomrule    
    \end{tabular}
    }
    \caption{Results of mean (i.i.d) and worst-group accuracy (WGA) and standard deviation over 10 random seeds on four image and text debiasing benchmarks. The proposed method, which does not require bias labels from the training (Tr.) and validation (Val) set, achieves superior or comparable WGA compared to methods like GroupDRO and DFR which require bias labels.}
    \label{tab:results-celeba}
\end{table*}

\begin{table}[!t]
\centering

\smallskip\noindent
\resizebox{0.7\columnwidth}{!}{%
\begin{tabular}{cccccc}
\toprule
 & \multicolumn{2}{c}{\textsc{Bias Label}} & \multicolumn{3}{c}{\textsc{Chexpert}} \\ 
 \cmidrule(r{4pt}){2-3} \cmidrule(r{4pt}){4-6}
 & \textsc{Train} & \textsc{Val} & \textsc{i.i.d} & \textsc{WGA} & \textsc{Gain} \\ \midrule
\textsc{ERM} & \xmark & \xmark & 89.78 & 31.21 & - \\
\midrule
\textsc{JTT\cite{liu2021just}} & \xmark & \xmark & 75.20 & 60.40 & +29.19 \\
\textsc{Ours} & \xmark & \xmark & \textbf{81.94} & \textbf{70.21} & \textbf{+40.00} \\ \midrule
\textsc{GDRO\cite{sagawa2019distributionally}} & \cmark & \cmark & 78.90 & 74.50 & +43.29 \\ \bottomrule
\end{tabular}%
}
\caption{Results on Chexpert Dataset. \label{tab:med}}
\end{table}

\subsection{Debiasing the Classifier} 
Even with transformed features, a classifier can still make false predictions due to high coefficients on spurious features. Here we train a new classifier on the balanced-sampled data aligning with the true data distribution.

\textbf{Minority Sampling} For better debiasing, the training data should be sampled fairly from the majority and minority groups, which are found in Section \ref{sec:susf}. 
Formally, let $\mathcal{D} = \{(\vec x_i, y_i)\}_{i=1}^{N}$ be the entire dataset, and $\mathcal{M}_1, \mathcal{M}_2 \subset \mathcal{D}$. 
\begin{equation}
    \mathcal{M}_1=\{(\vec x_{i_1}, y_{i_1}) \mid d(\vec x_i,\rho)>0 \land d(t(\vec x_i),0.5)<0\}
\end{equation}
\begin{equation}
    \mathcal{M}_2=\{(\vec x_{i_1}, y_{i_1}) \mid d(\vec x_i,\rho)<0\}
\end{equation}
The sampling process can be defined as \(\mathcal{S}_{M_i} = \{(\vec x_{1}, y_{1}),(\vec x_{2}, y_{2}), \dots, (\vec x_{B}, y_{B}) \mid (\vec x_{k}, y_{k}) \in \mathcal{M}_i, 1 \leq k \leq B\}\), where $\mathcal{S}_{M_1}$ and $\mathcal{S}_{M_2}$ are the sampled sets from $\mathcal{M}_1$ and $\mathcal{M}_2$, respectively. The new classifier \(h'\) is trained using Cross-Entropy (CE) loss, as follows:
\begin{equation}
\mathcal{L}_{cls} = \mathcal{L}_{CE}(h'(\mathbf{sg}[t(\vec x_i)]),y_i)+\mathcal{L}_{CE}(h'(\mathbf{sg}[t(\vec x'_i)]),y'_i),
\end{equation}

where $(\vec x_i,y_i)$ and $(\vec x'_i,y'_i)$ are the $i$-th example in the $\mathcal{S}_{M_1}$ and $\mathcal{S}_{M_2}$, and  \(\mathbf{sg[\cdot]}\) is the stop-gradient operation. We also conduct ablation of sampling in \cref{sec:abl}. 

\section{Experiments}

\subsection{Experiment Settings}

In this section, we provide a detailed description of the experimental setup.

\textbf{Datasets \& Metrics} To validate the proposed methods, we conduct experiments on both image and text tasks using five common benchmark datasets: The Waterbirds \cite{sagawa2019distributionally} dataset comprises 4,795 images, with class labels of landbirds and waterbirds, and a spurious attribute of background. CelebA \cite{liu2015deep} contains over 200,000 images, with hair color labeled and gender as a spurious attribute. MultiNLI \cite{williams2018broad} includes over 400,000 sentence pairs categorized into entailment, contradiction, or neutral, with the presence of negation words as a spurious attribute. CivilComments-WILDS \cite{koh2021wilds} also has over 400,000 text samples, labeled toxic or non-toxic, with spurious attributes related to demographic identities, including male, female, LGBTQ, Christian, Muslim, other religions, Black, and White. Finally, the CheXpert dataset includes over 200,000 chest radiographs labeled as ill or non-ill, without a specified spurious attribute. We report the WGA and mean accuracy on the test set using checkpoint from the last epoch.

\textbf{Network Architecture} We use ResNet-50 \cite{he2016deep} for image classification tasks, and BERT-base-uncased \cite{devlin2019bert} for text tasks as in most previous works.

\textbf{Implementation Details} All experiments were conducted on Nvidia GeForce 3090 GPU with 24GB VRAM using PyTorch \cite{paszke2019pytorch}. We first train an ERM model as in the DFR \cite{kirichenko2022last} study. For the CheXpert dataset, we follow the preprocessing and groups in the SubPopBench \cite{yang2023change} for a fair comparison. 
Then a transformation is learned using the pre-extracted embeddings from the ERM-trained model and data in the validation set. The AdamW optimizer is used with a learning rate of 1e-3 and zero weight decay.

\begin{figure*}[!t]
    \centering
    \includegraphics[width=\linewidth]{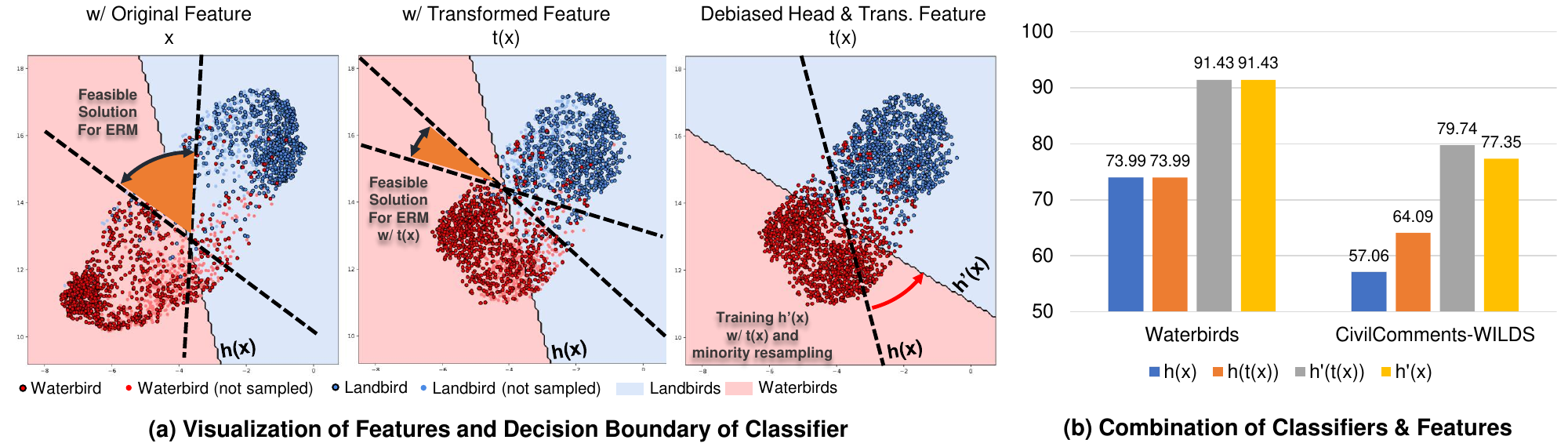}
    \caption{\textbf{(a)} Features and the decision boundary using models trained on the waterbirds dataset, using group-balanced sampling for better visualization.
    1) Original features $\vec x$ allows more biased solutions for the ERM training; 2) Elimination of spurious features using $t(\vec x)$ leaves a smaller room for biased solutions; 3) Finetuning $h'(\vec x)$ using $t(\vec x)$ results in a unbiased classifier. \textbf{(b)} Combinations of classifiers of ERM $h(\vec x)$ and debiased $h'(\vec x)$, and the features of raw $\vec x$ and transformed $t(\vec x)$. The debiased classifier $h'(\vec x)$ performs well using the original features $\vec x$ indicating $h'(\vec x)$ relies on core rather than spurious features.}
    \label{fig:ablumap}
\end{figure*}

\subsection{Comparison with the state-of-the-art Methods}

We compare the NSF against several state-of-the-art methods, including  JTT \cite{liu2021just}, MaskTune (MT) \cite{asgari2022masktune}, CNC \cite{zhang2022correct}, AFR \cite{qiu2023simple}, GroupDRO (GDRO) \cite{sagawa2019distributionally}, DFR \cite{kirichenko2022last} and uLA \cite{tsirigotis2024group}, and report mean (i.i.d) and worst-group accuracy (WGA) and standard deviation (std) over 10 random seeds on four image and text debiasing benchmarks.
As in \cref{tab:results-celeba},
the proposed method outperforms competing methods with an average of 82.01\% and low std.

\subsection{Mitigating Bias in Medical Domain}
We conduct experiments on the Chexpert \cite{irvin2019chexpert} dataset to validate the proposed method in the field of medical imaging, since unknown biases are fatal for automatic diagnostic, as in \cref{tab:med}. The proposed method shows a significant improvement compared to baseline methods.

\subsection{Ablation Study}
\label{sec:abl}
We conduct experiments on two debiasing benchmarks, the Waterbirds (by default) and CivilComments-WILDS to validate each component in the proposed NSF, and the worst group accuracy is presented in \cref{tab:abl} and \cref{fig:ablumap}(b). For a better understanding, we visualize the features and decision boundary in the waterbirds dataset using UMAP \cite{mcinnes2018umap}.

\textbf{The ERM \& Sampling Bias} As shown in \cref{fig:ablumap}(a), we plot group-balance sampled data from pre-extracted embeddings in the waterbird dataset, and borders are added as the ratio they are sampled in the training set for better visualization. It can be seen that such a sampling bias in the training data results in a larger feasible solution space for the ERM training, and most of them leverage the spurious feature for separating those classes, leading to a model bias. 

\begin{table}[!t]
\centering
\smallskip\noindent
\resizebox{0.7\columnwidth}{!}{%
\begin{tabular}{lccccc}
\toprule
 &   \multicolumn{2}{c}{\textsc{Feature}} &  \multicolumn{2}{c}{\textsc{Finetuning}} & \\ \cmidrule(r{4pt}){2-3} \cmidrule(r{4pt}){4-5}
 \textsc{Method}  & $\vec x$ & $t(\vec x)$ & \textsc{ERM} & \textsc{MR} & \textsc{WGA} \\ \midrule
\textsc{Baseline}   & \cmark &   & \cmark &&  72.60\\
\textsc{ERM}$+t(\vec x)$  &  & \cmark &  \cmark & &   87.85\\
MR$+\vec x$  & \cmark & &  &\cmark&86.92 \\
\textsc{Ours}   &  &\cmark & &\cmark &91.43\\
\bottomrule
\end{tabular}%
}

\caption{Ablation of components using the waterbirds dataset. Finetuning the classifier using the transformed features $t(\vec x)$ without resampling results in an increase of 15.25\% in WGA, proving that eliminating spurious features reduces the possibility of model bias. The similar good performance using minority sampling illustrates the correctness of the majority and minority groups found, and bias-sampled data makes a biased model.\label{tab:abl}}
\end{table}

\begin{table}[!t]
\centering
\smallskip\noindent
\resizebox{0.7\columnwidth}{!}{%
\begin{tabular}{lcccc}
\toprule
 &   \multicolumn{2}{c}{\textsc{Groups}} & \multicolumn{2}{c}{h(t($\vec x$))}  \\ \cmidrule(r{4pt}){2-3} \cmidrule(r{4pt}){4-5}
 \textsc{Method} & \textsc{Random}& $U_k, V_k$   & \textsc{i.i.d}& \textsc{WGA} \\ \midrule
\textsc{Baseline}  &  &  &  $97.30$ & $72.60$\\
\textsc{Random}  & \cmark & &90.60 & 73.99\\
$U_k, V_k$  & &\cmark & 95.65& 91.12\\
\bottomrule
\end{tabular}%
}
\caption{Ablation of the $U_k$ and $V_k$ used in learning $t(\vec x)$ using the waterbirds dataset. Replacing each sample in the $U_k$ and $V_k$ with a random one (equal to the sample mean) causes a significant drop in mean accuracy of 6\%, indicating that transforming to a biased mean corrupts the representation. Taking the example of dog vs. cat, it transforms a white cat to black (color of the majority group).\label{tab:abl1}}
\end{table}

\textbf{Ablation of Components} As in \cref{tab:abl}, both proposed components improve WGA, and the combination of them is the best, demonstrating their effectiveness.

\textbf{The Estimation of Bias-Invariant Feature} Replacing examples in the $U_k$ and $V_k$ with randomly selected examples used in learning $t(\vec x)$, as in \cref{tab:abl1} cause a drop of mean accuracy to 90.60\%, indicating that transforming to a biased mean corrupts the representation, highlighting the importance of using unbiased mean values for neutralizing the spurious features.

\textbf{Transformed Feature $t(\vec x)$} As shown in \cref{tab:abl}, finetuning the classifier with transformed features \( t(\vec x) \) improves WGA by 15.25\%, demonstrating that eliminating spurious features reduces model bias. \cref{fig:ablumap}(b) shows both \( h'(\vec x) \) and \( h'(t(\vec x)) \) perform well with the debiased classifier, with the transformed features achieving even better results, highlighting the effectiveness of eliminating spurious features.

\begin{figure}
    \centering
    \includegraphics[width=0.8\linewidth]{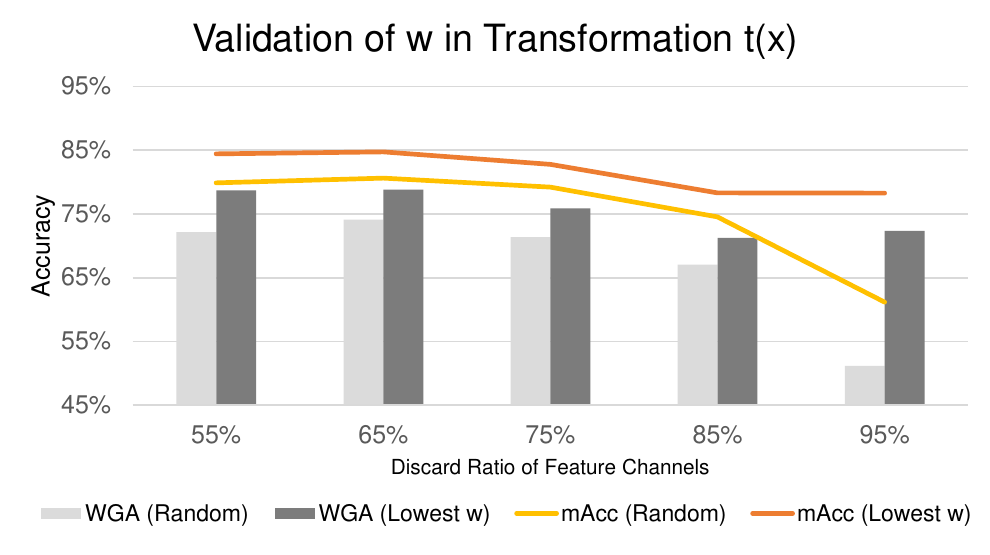}
    \caption{The WGA and mAcc of discarding channels by the lowest of the coefficient $w$ in the transformation $t(x)$ are higher than choosing randomly, validating lower $w$ highly correlate with spurious features so that they are eliminated, as in \cref{fig:tx}.}
    \label{fig:ablw}
\end{figure}

\textbf{Channels of Spurious Features} Discard some channels of feature by the lowest of the coefficient $w$ in the transformation $t(x)$ outperform random selection of same proportion. This validate low $w$ highly correlate with spurious features so that they are eliminated, as in \cref{fig:ablw}. 

\textbf{Debiased Classifier $h'(\vec x)$} To validate if the classifier successfully removes the bias, we test combinations of classifiers of ERM $h(\vec x)$ and debiased $h'(\vec x)$, and the features of raw $\vec x$ and transformed $t(\vec x)$, as in \cref{fig:ablumap}(b). The improvement of $h'(\vec x)$ using the original features $\vec x$ indicates the successful removal of dependence on spurious features.

\begin{figure}[!t]
    \centering
    \includegraphics[width=0.9\linewidth]{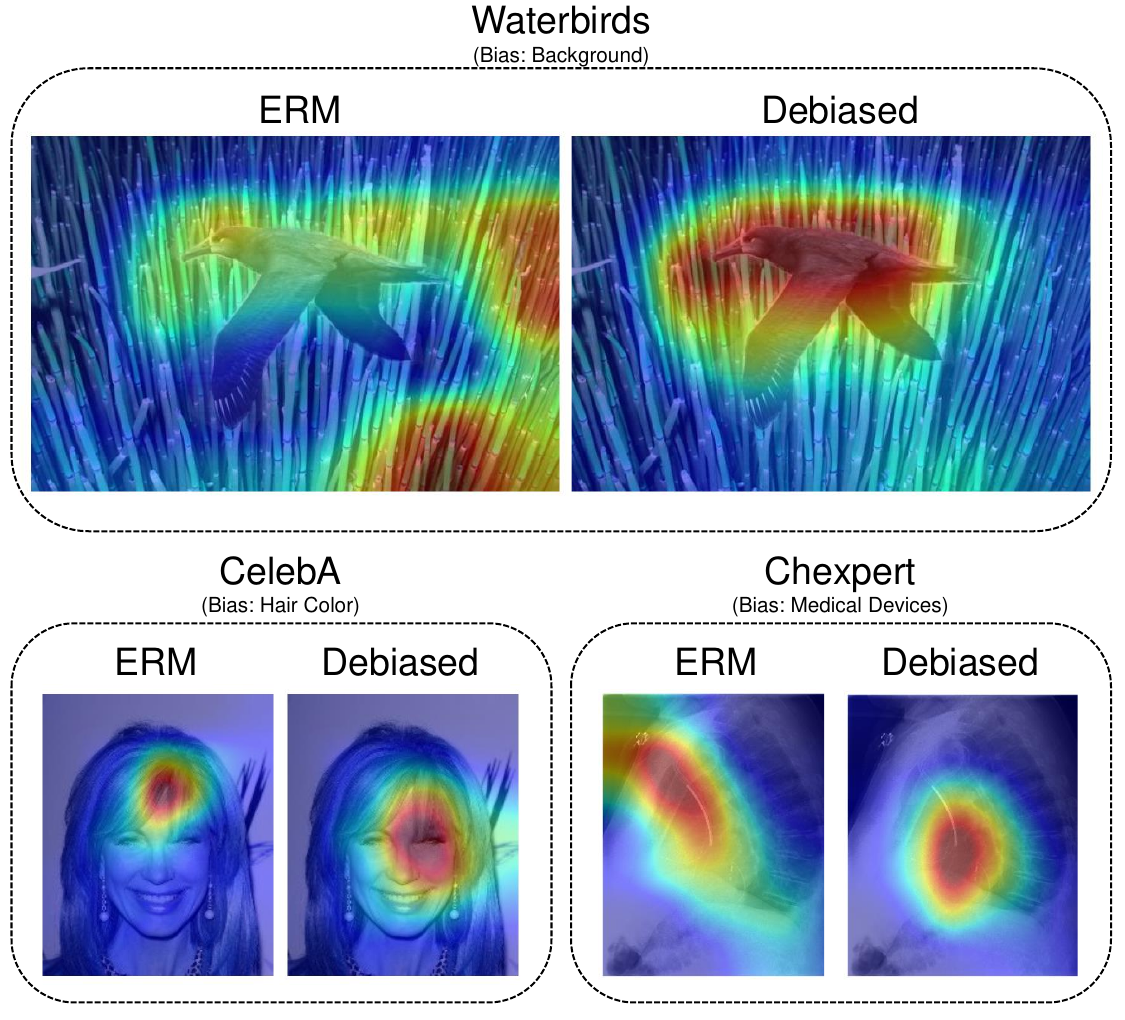}
    \caption{Biases found on the Waterbirds, CelebA, and CheXpert Datasets. The found biases are aligned with known biases (background, hair color, and medical devices).}
    \label{fig:cam}
\end{figure}

\section{Discussion}

\textbf{The Biases Found on Image Datasets} 
As the CAM \cite{selvaraju2017grad} in \cref{fig:cam}, for Waterbirds, the ERM model focuses on the background, while the debiased model centers on the bird's body. In CelebA, the ERM model highlights the hair, whereas the debiased model focuses on the face. In CheXpert, the ERM model targets medical devices, while the debiased model concentrates on clinically relevant areas. These visualizations show that the found biases are aligned with the known bias (the background, the hair color, and medical devices) in those datasets and debiasing leads to models using more relevant patterns.

\textbf{Results on Different Architecture}
ViT-s with ours achieves +5.56 in mAcc (90.78$\xrightarrow[]{}$96.34) and +21.96 in WGA (67.13$\xrightarrow[]{}$89.09) on Waterbirds compared to ERM.

\textbf{Swapping the Target $y$ and Spurious Attribute $a$} on CelebA results in +1.87 in WGA (90.56$\xrightarrow[]{}$92.43) and -2.79 in mAcc (98.60$\xrightarrow[]{}$95.81) compared to ERM.

\begin{figure}
    \centering
    \includegraphics[width=0.7\linewidth]{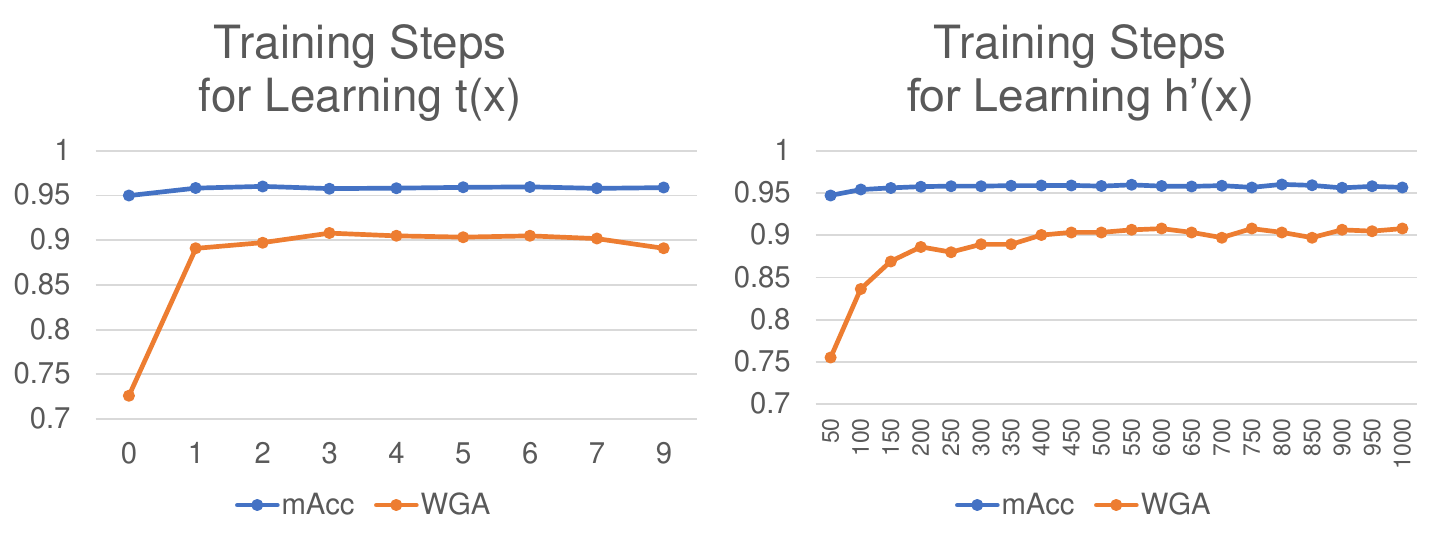}
    \caption{The impact of the training steps.}
    \label{fig:hsteps}
\end{figure}

\textbf{Training Efficiency \& Convergence} 
As shown in \cref{tab:spd}, it takes only a few seconds for the proposed method to remove the bias, highlighting advantages in training time and cost. 
We abbreviate WB for Waterbirds, CA for CelebA, MN for MultiNLI, and CC for CivilComments-WILDS. \cref{fig:hsteps} shows a clear trend of improvement in performance with increased training then begins to plateau, indicating a successful convergence.

\begin{table}[!t]
\centering
\smallskip\noindent
\resizebox{0.7\columnwidth}{!}{%
\begin{tabular}{lccccc}
\toprule
 \textsc{Comp.} &   {\textsc{WB}} & {\textsc{CA}} & {\textsc{MN}} & {\textsc{CC}} & \textsc{Mean}  \\ \midrule
\textsc{$t(\vec x)$} &1s&2s&15s&2s&5s\\
\textsc{$h'(\vec x)$} &1s&1s&1s&1s&1s\\
\bottomrule
\end{tabular}%
}
\caption{Comparison of training time on different datasets.\label{tab:spd}}
\end{table}

\textbf{Limitations} The NSF follows findings in \cite{hermannfoundations}, which suggest models favor strong features even when less predictive. This assumption may limit the applicability of NSF under weaker biases, which we should address in future works. 

\section{Conclusion}

This work introduces NSF, a novel method for mitigating unknown biases in DNNs. NSF effectively debias models by identifying and eliminating spurious features, while reducing their influence in the classifier, using insights from the model’s bias-fitting mechanism. Extensive experiments across multiple benchmarks show that NSF achieves state-of-the-art results in both vision and text tasks, with minimal computational cost, demonstrating its efficiency. Due to its easy integration, non-intrusive nature, and high efficiency, NSF provides a new choice for addressing unknown biases in DNNs.

{
    \small
    \bibliographystyle{ieeenat_fullname}
    \bibliography{main}
}

\end{document}